\documentclass[letterpaper, 10 pt, conference]{ieeeconf}  % Comment this line out if you need a4paper

\IEEEoverridecommandlockouts                              % This command is only needed if 
\usepackage{xspace}
\usepackage{float}
\usepackage{url}
\usepackage{hyperref}
\hypersetup{
    colorlinks=true,       % Enable colored links
    linkcolor=magenta,     % Color for internal links
    urlcolor=magenta       % Color for URLs
}
\newcommand{\algabbr}{POGS\xspace}

\usepackage[tracking=true]{microtype}
\usepackage{amsmath} % assumes amsmath package installed
\usepackage{amssymb}  % assumes amsmath package installed
\usepackage{graphicx}
\usepackage{epsfig} % for postscript graphics files
\usepackage{mathptmx} % assumes new font selection scheme installed
\usepackage{times} % assumes new font selection scheme installed
\usepackage{amsmath} % assumes amsmath package installed
\usepackage{amssymb}  % assumes amsmath package installed
\usepackage{xspace}

\usepackage{listings}
\usepackage{float}

\usepackage{afterpage}
\usepackage{url}

\usepackage[dvipsnames]{xcolor}
\usepackage[font={footnotesize}]{caption}

\usepackage{booktabs} % for professional tables
\usepackage{tikz}
\usetikzlibrary{positioning,shapes.geometric,arrows,fit,shapes.symbols,svg.path,tikzmark,calc}
\usepackage{textcomp}
\usepackage{subcaption}
\pgfdeclarelayer{background}
\pgfdeclarelayer{foreground}
\pgfsetlayers{background,main,foreground}
\usepackage{siunitx}
\usepackage{adjustbox}
\usepackage{array}
\usepackage{multirow}
\usepackage{siunitx}
\usepackage{gensymb}

\let\labelindent\relax %https://tex.stackexchange.com/questions/170772/command-labelindent-already-defined

\usepackage[inline]{enumitem} % for inline lists, e.g. \begin{enumerate*}[label=\roman*)]  See https://tex.stackexchange.com/questions/146306/how-to-make-horizontal-lists

\usepackage[backend=biber,
            hyperref=true,
            url=true,
            isbn=false,
            doi=false,
            backref=false,
            style=ieee,
            natbib=true,%compatibility aliases
            mincitenames=1,
            maxcitenames=1,
            citestyle=numeric-comp,
            sorting=none,
            block=none,
            maxbibnames=99]{biblatex}
\renewcommand{\bibfont}{\small}
\title{\LARGE \bf
Persistent Object Gaussian Splat (POGS) for Tracking \\Human and Robot Manipulation of Irregularly Shaped Objects
}

\author{Justin Yu$^{*1}$, Kush Hari$^{*1}$, Karim El-Refai$^{*1}$, Arnav Dalal$^{1}$, Justin Kerr$^{1}$, Chung Min Kim$^{1}$, \\ Richard Cheng$^{2}$, Muhammad Zubair Irshad$^{2}$, Ken Goldberg$^{1}$ % <-this % stops a space
\thanks{$^{*}$ Equal contribution}%
\thanks{$^{1}$The AUTOLab at UC Berkeley (automation.berkeley.edu).}
\vspace{0.4cm} \\
\url{https://berkeleyautomation.github.io/POGS}
\vspace{-0.2cm}
\thanks{$^{2}$Toyota Research Institute, Los Altos, CA.}
}

\begin{document}

\maketitle
\thispagestyle{empty}
\pagestyle{empty}

%%%%%%%%%%%%%%%%%%%%%%%%%%%%%%%%%%%%%%%%%%%%%%%%%%%%%%%%%%%%%%%%%%%%%%%%%%%%%%%%
\begin{abstract}
Tracking and manipulating irregularly-shaped, previously unseen objects in dynamic environments is important for robotic applications in manufacturing, assembly, and logistics. Recently introduced Gaussian Splats \cite{kerbl3Dgaussians} efficiently model object geometry, but lack persistent state estimation for task-oriented manipulation. We present Persistent Object Gaussian Splat (POGS), a system that embeds semantics, self-supervised visual features, and object grouping features into a compact representation that can be continuously updated to estimate the pose of scanned objects. \algabbr updates object states without requiring expensive rescanning or prior CAD models of objects. After an initial multi-view scene capture and training phase, \algabbr uses a single stereo camera to integrate depth estimates along with self-supervised vision encoder features for object pose estimation. POGS supports grasping, reorientation, and natural language-driven manipulation by refining object pose estimates, facilitating sequential object reset operations with human-induced object perturbations and tool servoing, where robots recover tool pose despite tool perturbations of up to 30°. POGS achieves up to 12 consecutive successful object resets and recovers from 80\% of in-grasp tool perturbations.

\end{abstract}

%%%%%%%%%%%%%%%%%%%%%%%%%%%%%%%%%%%%%%%%%%%%%%%%%%%%%%%%%%%%%%%%%%%%%%%%%%%%%%%%

% Two or three meaningful keywords should be added here
% \keywords{state estimation, interactive perception}

%===============================================================================

\section{Introduction} \label{sec:intro}

In environments like factories, workshops, or homes, robots must not only successfully identify and manipulate objects but also adapt to changes in object pose over time. 
The challenge is greater when dealing with irregularly shaped objects for which obtaining an accurate Computer-Aided Design (CAD) model is impractical. While any physical object can in principle be CAD-modeled, this process is often labor-intensive and may require reaching out to manufacturers or purchasing specialized scanning equipment. Approaches that rely on predefined CAD models struggle in scenarios where such models are unavailable, limiting adaptability to previously unseen objects \cite{zhang2019tracking, wuest2007tracking, wiedemann2008recognition}. Traditional and deep RGBD or point cloud object tracking methods are attractive as components of state estimators for robotic manipulation because they do not require predefined meshes or CAD models \cite{zhou2024survey, zheng2017online}. However, many of these approaches fail to effectively integrate geometric information across multiple object viewpoints or timesteps, and do not address the estimation or reconstruction of occluded object regions based on prior information. As a result, they struggle to maintain a persistent and holistic object representation over time.
% The challenge is greater when dealing with irregularly shaped objects that are not easily modeled with Computer Aided Design (CAD) software. Approaches that rely on predefined CAD models struggle to adapt to unseen objects 

\begin{figure}
    % \begin{minipage}{\textwidth}
    \centering
    \includegraphics[width=\linewidth]
     {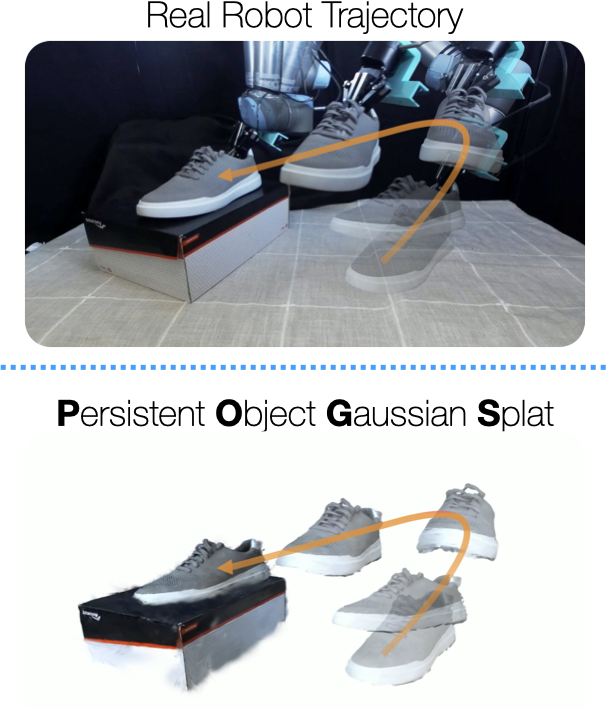}
    \caption{\textbf{Autonomous Object Manipulation and Tracking with POGS Unified Representation} (Top) A robot autonomously performs a pick and place primitive to move the shoe onto a shoebox given input natural language pick query "shoe" and place query "shoebox". (Bottom) A POGS unified representation enables language querying, grasp sampling, and continuous tracking of irregular objects as they move.}
    \label{fig:splash}
    \centering
    \vspace*{-0.3in}
\end{figure}

% It is also valuable to represent objects such that they can be queried with natural language.
Implicit 3D representations like NeRFs \cite{mildenhall2021nerf} offer high-quality scene reconstructions but are ill-suited as a representation for dynamic scenes where objects may be moved and re-oriented. Recently, Gaussian Splatting \cite{kerbl3Dgaussians} was introduced to create high quality 3D reconstructions by explicitly modeling scenes as a set of 3D gaussians that can be partitioned to allow rigid transforms on object-level clusters. 
% Along with being capable of tracking objects over time, it is desirable for a scene representation to be queryable with natural language and have accurate 3D object segmentation.

\begin{figure*}[ht]
    \centering
    \includegraphics[width=0.9\linewidth]{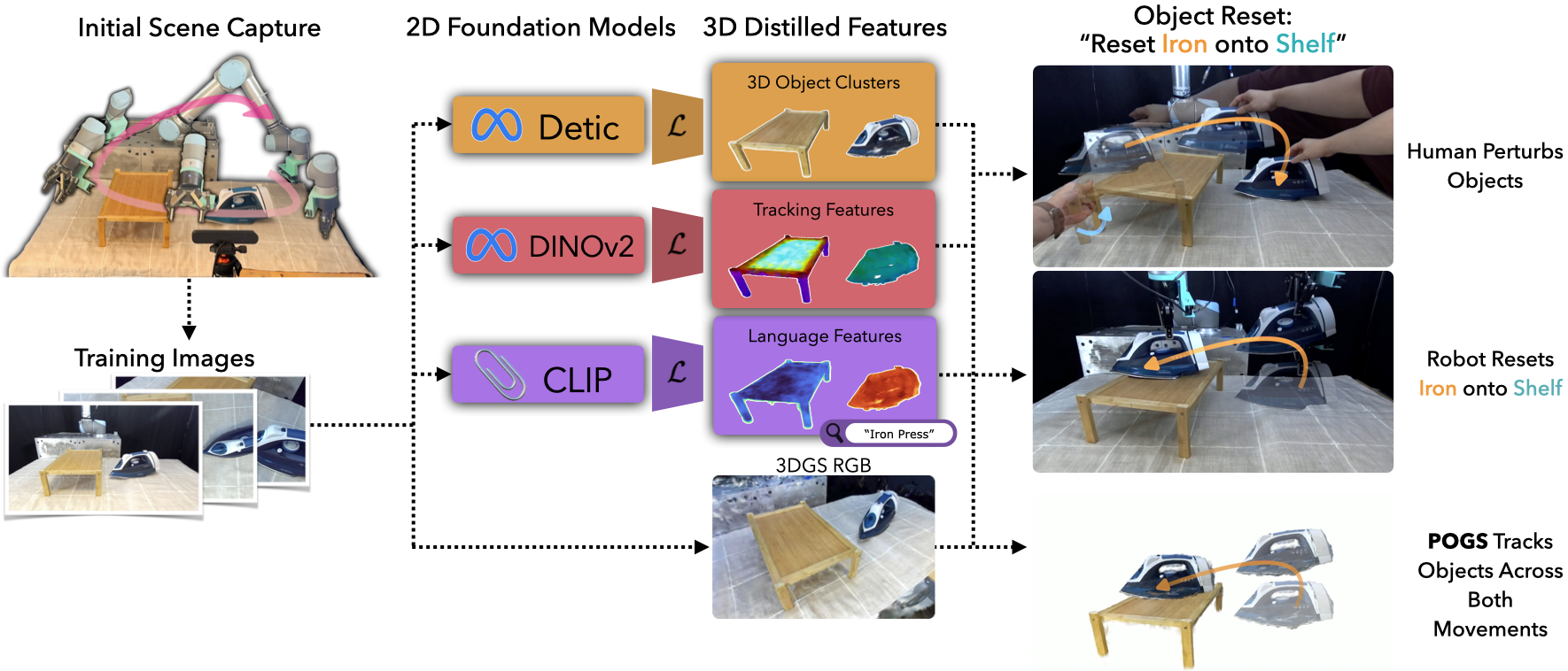} 
    \caption{\textbf{\algabbr Pipeline} After capturing multiple images of a scene using a robot wrist-mounted ZED mini, \algabbr segments objects using Detic, extracts DINO features, and embeds language through CLIP. Training images are used to optimize a 3DGS, and features extracted from 2D foundation models are distilled into feature fields, producing our \algabbr unified representation. During robot object reset and tool servoing, the \algabbr is updated based on depth geometry and DINO tracking features.}
    \label{fig:enter-label}
    % \vspace{-2in}
    \vspace{-0.5em}
\end{figure*}

To enable online state estimation, tracking, and manipulation of unseen objects in dynamic environments, we present Persistent Object Gaussian Splat (\algabbr), an editable object-centric feature field representation with embedded language features, self-supervised visual features, and object-level grouping features to support robot manipulation. \algabbr uses 3D Gaussian Splatting (3DGS) to model full 3D geometry of irregular objects, allowing for continuous updates as the scene evolves. 
% Unlike methods that require pre-existing CAD models, \algabbr performs an initial training phase to analyze a provided scene and explicitly represent the geometric and image features of unknown objects.

By embedding features from encoders and detectors pre-trained on internet-scale datasets such as CLIP \cite{radford2021learning}, DINO \cite{oquab2023dinov2}, and Detic \cite{zhou2022detecting}, \algabbr can respond to open-vocabulary natural language queries and also identify, track, and manipulate objects even without any predefined models. As such objects are moved by humans or robots, \algabbr can update their state online, allowing for flexible, multi-step tasks that require continuous interaction with dynamic objects, eliminating the need to re-scan the environment.
This paper makes the following contributions:
\begin{itemize}

\item Persistent Object Gaussian Splat (\algabbr), a novel feature field representation for tracking and manipulating previously unseen irregularly shaped objects.
\item A robot system for creating and using \algabbr to perform object-reset and tool servoing tasks.
\item Physical robot experiments on object reset with an average pose error of 2.92 cm. 
\item Physical robot experiments for tool servoing where targets are moved up to 30$\degree$ and the tool can recover from human perturbations 80\% of the time.

\end{itemize}
\section{Related Work}

% \begin{figure*}[!t]
% \centering
% \includegraphics[width=0.8\linewidth]{images/ex_method.png}
% \caption{}
% \label{fig:method}
% \vspace*{-0.3in}
% \end{figure*}

\subsection{Feature Fields for Robotics}

Recent advances in foundation vision models such as CLIP \cite{radford2021learning} and DINO \cite{oquab2023dinov2} have enabled many methods that perform robot manipulation from visual features. Works \cite{huang2023voxposer, liu2024ok, chen2023open, huang23vlmaps, conceptfusion, yokoyama2024vlfm, gu2024conceptgraphs} have used CLIP with point-based fusion methods to build open-vocabulary 3D representations. However, visual occlusion and multi-view pose misalignment can hinder consistent semantic fusion across the 3D scene. 

To address this, more recent approaches such as DFF \cite{kobayashi2022distilledfeaturefields} and LERF \cite{kerr2023lerf} have proposed distilling learned features into neural radiance fields (NeRFs) \cite{mildenhall2021nerf} by aggregating information across multiple views and scales. F3RM \cite{shen2023F3RM} and LERF-TOGO \cite{lerftogo2023} extend these works respectively for robot manipulation. However, NeRF-based representations are limited by NeRF's training speed and implicit spatial representation, making it impossible to update when objects move without further scene-scale optimization. Works like Dex-NeRF \cite{ichnowski2022dex} and Evo-nerf \cite{kerr2022evo} attempt to address this by partially re-scanning scenes to account for object movement; however, this process remains computationally intensive, limiting its suitability for online updates.

An alternative to NeRF is Gaussian Splatting \cite{kerbl3Dgaussians}, which models scenes using explicit 3D Gaussian primitives, enabling faster training and rendering while maintaining high fidelity in 3D reconstructions. Recent works \cite{legs2024iros, zhou2024feature, qiu-2024-featuresplatting, qin2024langsplat} have shown that Gaussian Splatting can also integrate semantic and grouping features. GaussianGrasper \cite{zheng2024gaussiangrasper} extends these approaches to robotic manipulation by updating the scene representation after objects are moved, using the robot end-effector pick-and-place transform followed by a few views to fine-tune the Gaussian Splat. In this work, we develop a method capable of updating the scene where a human can also move the objects repeatedly without any partial re-scans of the scene.

\subsection{Object Tracking for Manipulation}
Object pose estimation networks \cite{wen2021bundletrack, wen2023bundlesdf, wen2024foundationpose, sun2022onepose} are able to track the 6DOF pose of an object of interest, but typically not multiple objects at once in a scene without scaling compute requirements. Using Gaussian Splatting, several works \cite{luiten2023dynamic, Wu_2024_CVPR, yang2023deformable3dgs} collect image data from one or multiple views over time for rendering dynamic scenes. However, these works focus on offline processing and pose interpolation rather than tracking and estimating object states online for manipulation tasks. Keypoint-based approaches \cite{florence2018dense, gao2021kpam, simeonov2022neural, yen2022nerfsupervision} model multiple objects in a scene as a set of keypoints. However, these methods are prone to tracking errors when objects rotate and keypoints become occluded. Some approaches \cite{huang2024rekep, wang2023d3fields} use multi-camera setups to help mitigate these issues. Our approach aims to achieve robust online object tracking and scene updating with a single stereo camera.

\subsection{Editable 3D Feature Fields}
Concurrently, other works develop editable 3D feature fields. GaussianGrasper \cite{zheng2024gaussiangrasper} and Splat-Mover \cite{shorinwa2024textbf}, allow Gaussian splat updates based on known robot end-effector movements. However, they assume robot-only object interactions whereas \algabbr can also track human object interactions. Object-Centric Gaussian Splats \cite{li2024object} and GraspSplats \cite{ji2024graspsplats} improve tracking but rely on static backgrounds or multi-camera systems. Robot See Robot Do \cite{kerrrobot} tracks part-level objects using monocular video, though only in an offline processing setting for zero-shot motion planning robot imitation from human demonstration. We extend this work to support online rigid multi-object tracking along with the aforementioned semantic and object-centric feature fields to create a unified 3D scene representation for zero-shot robotic manipulation.

\section{Problem Statement}

We consider a tabletop setting with irregular objects, defined as objects for which we do not have a detailed geometric (CAD) model. Given a single stereo camera, the objective is to track the 6D pose of each object over time and update the 3D scene models.
We make the following assumptions:
\begin{enumerate}
    \item Each object is rigid, simplifying the tracking to estimate rigid transforms from RGBD frames.
    \item There exists an initial scanning phase in which all objects are static. However, at the start of tracking, objects can be in different poses within 90\degree and 25 cm of their initial scan pose.
    \item All object surfaces are represented in \algabbr training views--with the exception of object surfaces in contact with the tabletop.
    \item The scene is well-lit with approximately uniform lighting and minimal shadowing.
    \item During object tracking, objects are not placed in configurations where they fully occlude each other.
    \item Object surfaces exhibit low specularity for more robust geometry reconstruction and visual feature extraction.
    
\end{enumerate}

We evaluate POGS with 2 types of robot experiments: object-reset and tool servoing. 

The goal of the object-reset experiment is to use natural language to query for an object to grasp and another query for where the grasped object will be placed. After each object reset, a human will randomly reconfigure both objects to different poses and the process is repeated until failure. We evaluate this experiment by recording the maximum number of sequential object resets before failure, the object grasp rate, the object place rate, and the object translation error in placement.

In tool servoing, the objective is for the robot to continuously align a grasped tool with a target, even as a human operator moves the target and alters the tool’s orientation within the robot’s grip. We evaluate this experiment by recording the success rate and average time taken to recover from in-grasp tool perturbations.

\section{Method}

% \subsection{Overview}
The \algabbr system has 3 phases: 
\begin{enumerate}
    \item \textbf{Scene Capture} phase to obtain a set of images of a novel environment in a multi-view manner, maximizing different perspectives on objects of interest.
    \item \textbf{Training} phase for aggregating and distilling information from captured views into a unified \algabbr representation. 
    \item \textbf{Persistent Object Tracking} phase for online tracking and updating of object poses as they move through the workspace from human or robot manipulation.
\end{enumerate}

\subsection{Scene Capture}

The initial scene is scanned using an RGBD ZED Mini stereo camera mounted on a UR5 robot end effector. We capture images from 35 views as the robot moves along a predefined trajectory around the workspace to have sufficient viewpoints. Depth images are obtained from stereo pairs with RAFT-Stereo inference \cite{lipson2021raft}, which are deprojected into a fused pointcloud and used for initializing the Gaussian Splat. We run DBSCAN~\cite{ester1996density} on the fused pointcloud to filter noise and floater points.

\subsection{Gaussian Splatting with Feature Fields}
The goal of the training phase is to fuse information from the collected multi-view images and generate the unified 3D representation \algabbr for all objects in the scene. Rendering and supervising color for \algabbr remains exactly the same as 3DGS. We additionally regularize the 3D geometry with a depth reconstruction objective to encourage Gaussian means to be positioned on object surfaces \cite{chung2024depth}. We employ feature rendering techniques \cite{zhou2024feature,shen2023F3RM} to simultaneously distill useful latent and explicit features from the 2D images into the 3DGS. In particular, we supervise into the 3DGS feature field: 
% 1) 2D Masks for singulating objects from the environment, 2) Multi-scale CLIP pyramid features for natural-language alignment, using the same method as \textcolor{red}{todo: cite LERF, LEGS}, and 3) DINO features for dense image-feature optimization based tracking, similar to the technique used in \textcolor{red}{todo: cite RSRD}. 
% \subsection{Object Masks}

\textbf{Grouping Features:} 2D object-level masks are supervised into 3D features for object clustering and singulating them from the environment.
We obtain 2D masks from the training images using the object detection and segmentation network Detic \cite{zhou2022detecting}.
% , though we note that any automatic mask generator or object segmentation method that produces reliable object-level 2D masks can serve this purpose. 

To distill 2D object masks into 3D gaussian partitions, we borrow principles from \cite{garfield2024,bhalgat2023contrastivelift3dobject} and train a feature embedding encoder $F_{\text{emb}}$ that passes an input gaussian mean position $\vec{x}\in \mathbb{R}^3$ through a hash-grid encoder~\cite{muller2022instant} followed by an MLP, outputting a $D$-dimensional embedding vector. 3D gaussian features are rendered from a specific camera location to produce a feature map. For this we use Nerfstudio's~\cite{tancik2023nerfstudio,ye2024gsplatopensourcelibrarygaussian} 3DGS tile-based rasterizer implementation, with gradients backpropagated through the MLP within $F_{\text{emb}}$.
3D grouping features are then supervised with the contrastive objective from~\citet{bhalgat2023contrastivelift3dobject}, which operates through two complementary mechanisms: (1) attracting features that belong to the same object mask by minimizing their distance in embedding space, and (2) repelling features from different object masks by maximizing their embedding distances.
We observe that computing and including a negative mask (wherever an object mask does not exist) is helpful in reducing group feature noise for the scene background (anything in the scene that is not a tracked object).

Before the tracking phase begins, the system must identify and segment individual objects within the scene. As proposed in GARField~\cite{garfield2024}, this is accomplished by clustering the group features using HDBSCAN. The result is a mapping from each 3D gaussian to a mask and label.
% Let:

% \begin{itemize}
%     \item $F_{\text{emb}}(x)$: Embedding vector for image-pixel $x\in\mathbb{R}^2$.
%     \item $\mathcal{M} = \{ M_1, M_2, \dots, M_N \}$: Set of $N$ instance masks in the batch.
%     \item $\mu_i$: Mean embedding vector for mask $M_i$, defined as
%     \begin{equation}
%     \mu_i = \frac{1}{|M_i|} \sum_{x \in M_i} F_{\text{emb}}(x)
%     \end{equation}
% \end{itemize}

% For randomly selected pairs of masks $(M_i, M_j)$ where $i \neq j$ and both masks meet the minimum size criteria, the inter-mask loss is defined as:

% \begin{equation}
% L_{\text{inter}} = \sum_{(i,j) \in \mathcal{S}} \text{ReLU} \left( m - \left\| \mu_i - \mu_j \right\|_2 \right)
% \end{equation}

% where $\mathcal{S}$ is the set of valid mask pairs and $m\in\mathbb{R}$ is the margin, representing the maximum allowable distance between individual embeddings.

% For each mask $M_i$, the intra-mask loss is:

% \begin{equation}
% L_{\text{intra}} = \sum_{i=1}^{N} \frac{1}{|M_i|} \sum_{x \in M_i} \text{ReLU} \left( \left\| F_{\text{emb}}(x) - \mu_i \right\|_2 \right)
% \end{equation}

% The overall grouping loss is computed by averaging over the total count of valid terms:

% \begin{equation}
% L_{\text{instance}} = \frac{1}{K} \left( L_{\text{inter}} + L_{\text{intra}} \right)
% \end{equation}

% where $K$ equals the total number of valid terms contributing to the loss (i.e., the count of mask pairs in $L_{\text{inter}}$ plus the count of masks in $L_{\text{intra}}$).

\textbf{Language Features:} To facilitate natural language queries in 3D, we incorporate multi-scale CLIP pyramid features distilled into the 3D gaussians, following the methodology described in~\cite{kerr2023lerf} and \cite{legs2024iros}. Specifically, a scale-conditioned language feature embedding function is defined, $F_{lang}(\vec{x}, s): (\mathbb{R}^3,\mathbb{R}) \rightarrow \mathbb{R}^D$ which maps a position $\vec{x}$ and physical scale $s$ to a language-aligned embedding vector. 

% We extract language-aligned features from each input image by passing multi-scale image pyramid crops through the CLIP image encoder. This multi-scale approach is crucial, as demonstrated in LERF \cite{kerr2023lerf}, for capturing semantic understanding across varying object sizes. 

As in LERF, during deployment we use the CLIP text encoder to obtain embedding vectors for arbitrary natural language input queries. The relevancy of each gaussian to a given text query is computed by taking the cosine similarity score between the gaussian's language embedding and the text embedding. 

\textbf{Self-Supervised Features for Object Tracking:}
We distill dense visual features extracted from DINOv2~\cite{oquab2023dinov2} into the 3D gaussians during training. These features are then supervised into the gaussians, enabling the model to render them at deployment time for optimizing object tracking objectives, similar to the method described in Robot See, Robot Do~\cite{kerr2024robot} and further detailed in the next sections.

Unlike the object grouping features and language features where we learn embedding functions to map inputs into feature space, the supervision of DINO visual features into \algabbr instead directly renders and optimizes trainable feature vectors of dimension-$d$ with each Gaussian primitive. To integrate the DINO features efficiently, we apply principal component analysis (PCA) to reduce their dimensionality from several hundred to $d=64$ dimensions. Without dimensionality reduction, storing per-Gaussian feature vectors would be computationally prohibitive.

We use Nerfstudio’s \cite{nerfstudio}
Splatfacto implementation of Gaussian Splatting with the gsplat~\cite{ye2024gsplatopensourcelibrarygaussian} backend and modify it with the aforementioned image encoders and feature supervision losses.

\subsection{Persistent Object Representation}
Because \algabbr contains language, grouping, and visual features in a single representation, \algabbr can be used to query for objects with natural language and track those objects online by representing each object as a cluster of gaussian primitives in 3D. For each gaussian cluster, the system uses the centroid of gaussians within that cluster as the object frame, canonicalized such that the initial pose of each object is rotated to align with the world frame.

\algabbr extracts DINO visual features from live stereo camera observations, from only the left camera of the stereo pair. Feature rendering from the \algabbr model can directly obtain a synthetic view of the object feature maps from the same perspective as the real camera by using calibrated extrinsics. Simultaneously, \algabbr captures depth maps that serve as ground truth geometry to further regularize object pose. To generate accurate depth maps from stereo images, we employ a neural depth estimation model developed by Toyota Research Institute (TRI)~\cite{shankar2022learned}, chosen for memory efficiency and real-time inference. This model operates effectively at an image resolution of 1080p, with depth inference frequency at approximately 30 Hz.

\begin{figure}[h]
    \centering
    \includegraphics[width=0.95\linewidth]{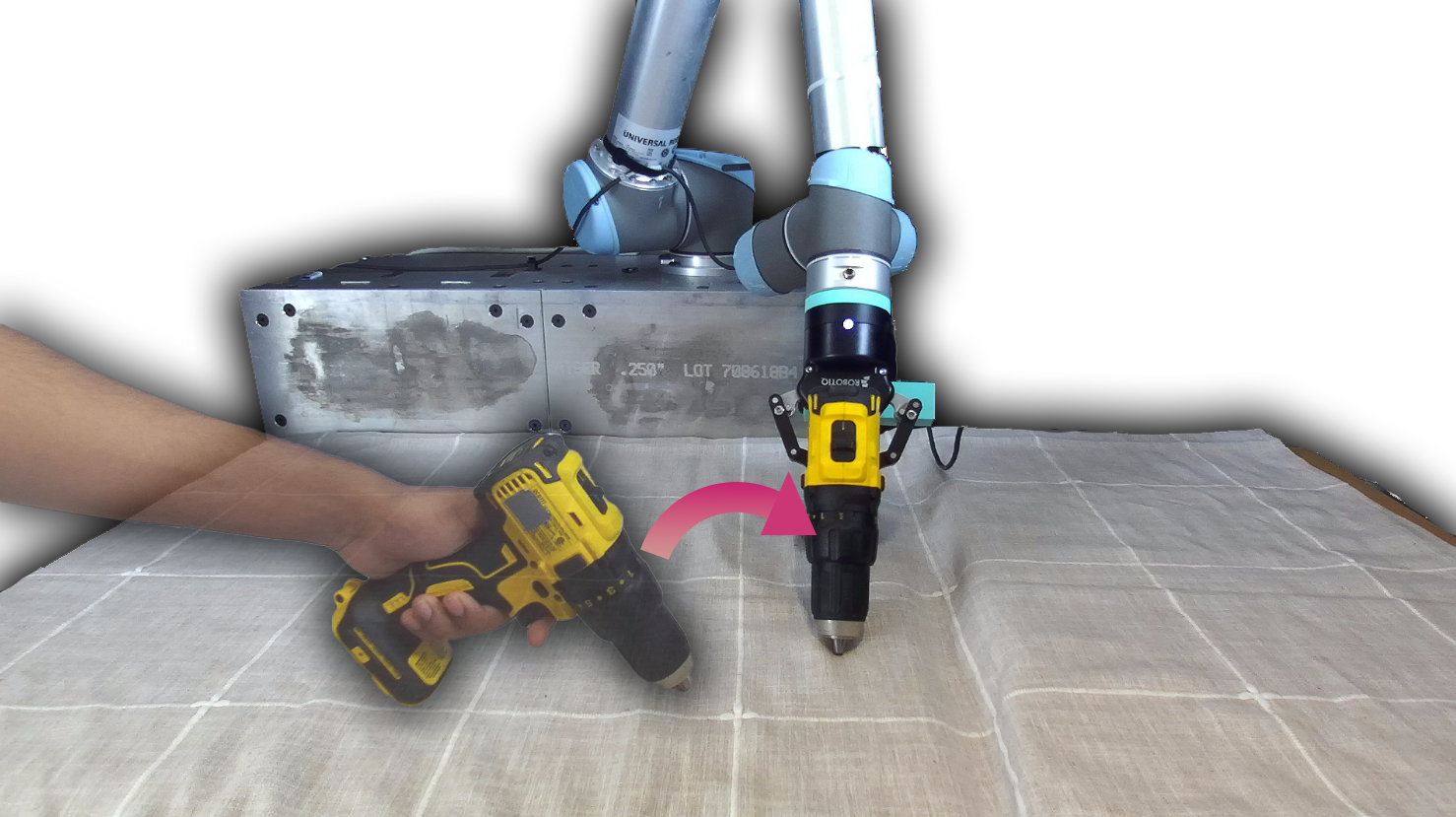}
    \caption{\textbf{Occluded Grasp Sampling} \algabbr is capable of sampling and performing robot grasps on geometry that is fully occluded from the observation camera view (shown). The drill handle is fully occluded by the motor body, yet our \algabbr unified representation enables handle grasping based on previously observed geometry.}
    \label{fig:drill_occluded}
    \vspace*{-0.2in}
\end{figure}
\subsection{Tracking with POGS}

Inspired by RSRD~\cite{kerr2024robot}, the core of the tracking algorithm is the computation of the loss between the distilled DINO features of the rendered Gaussian Splat and the observed images. This feature loss measures how well the current pose estimates visually align the rendered model with the actual objects. \algabbr also includes a depth loss term that compares the rendered depth maps with depth maps extracted from the real observations, enforcing geometric consistency.
The total loss is a weighted sum of the feature loss and depth loss, guiding the optimization to adjust per-object pose parameters until convergence.
Each Gaussian cluster pose parameter is optimized independently, allowing \algabbr to track multiple moving objects, without imposing constraints on their relative movements. unlike prior work in real-time tracking of gaussian splats.

% \subsection{Iterative Tracking}
For each new frame captured by the camera, \algabbr repeats the rendering, feature extraction, loss computation, and optimization steps. This iterative process continually refines the pose estimates, improving alignment between the rendered clusters and the observed images over time.

\subsection{Human \& Robot Manipulation}
We deploy \algabbr for tracking human and robot manipulation tasks where objects may be in varying poses compared to their initial positions in the scene capture. To facilitate robot grasping based on language queries, \algabbr first identifies the object cluster that corresponds to the query. The Gaussian means representing that object cluster are passed as a point cloud to Contact-GraspNet \cite{sundermeyer2021contact}, which generates potential grasp candidates along with their respective scores, and the highest scored grasp is executed. By using the Gaussian means, the object grasp is based on the full 3D object geometry embedded in \algabbr, which can be beneficial compared to methods that solely deproject depth and are partially occluded as seen in Figure \ref{fig:drill_occluded}.  
\begin{figure}[t]
    \centering
    \includegraphics[width=0.95\linewidth]{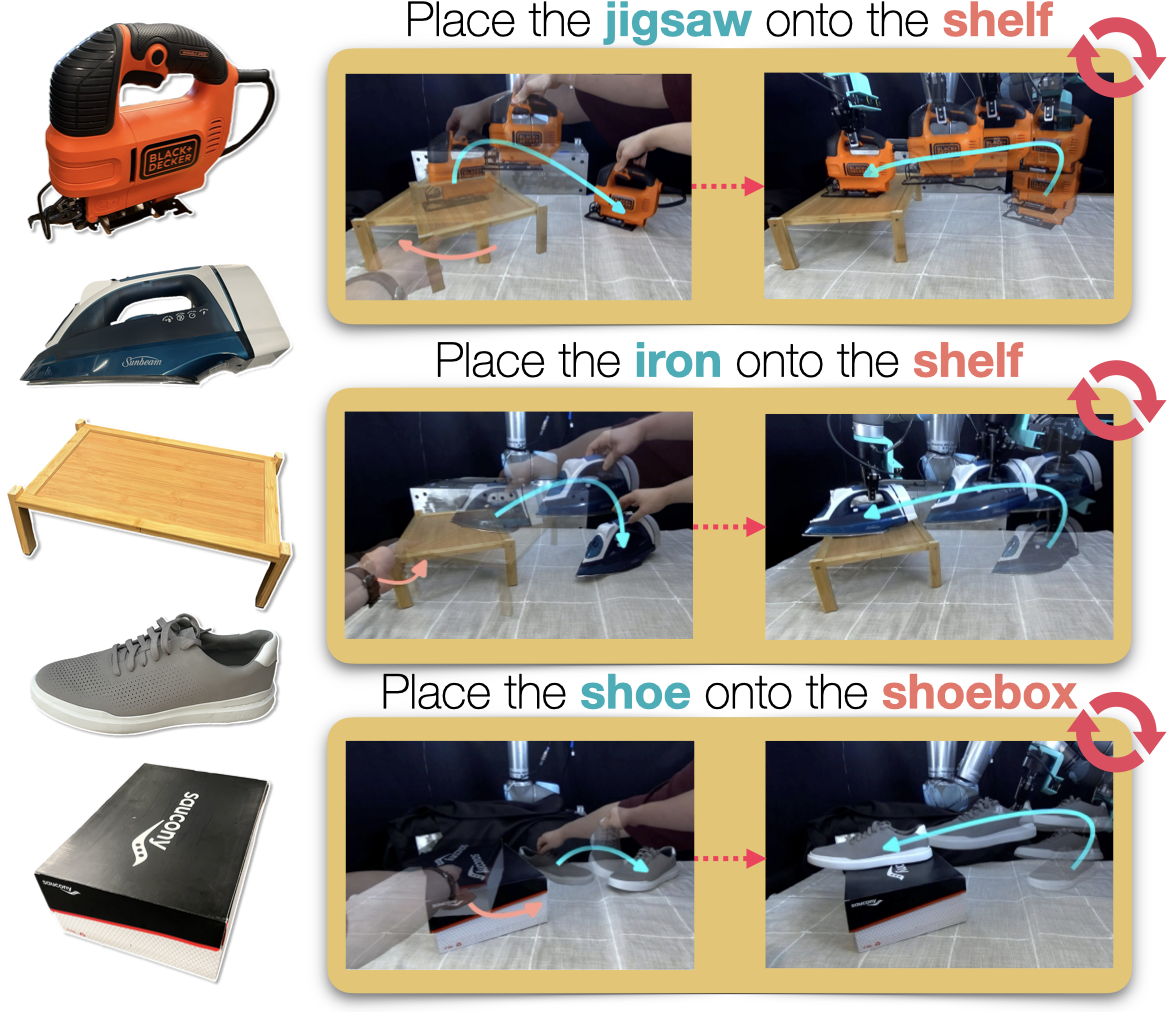}
    \caption{\textbf{Object Reset Experimental Setup} \textit{Middle:} A human randomly perturbs the configuration of the tracked objects according to the two tiers. \textit{Right:} A robot arm then plans a grasp on language-queried objects and performs object reset. This process repeats until errors in object state estimation are too high to
recover for grasping.}
    \label{fig:sequential}
    \vspace*{-0.2in}
\end{figure}

\setlength{\tabcolsep}{4pt}
\begin{table*}
\centering
\vspace{0.5em}
{\scriptsize
\begin{tabular}{lccc|c!{\vrule width 1.2pt}ccc|c!{\vrule width 1.2pt}ccc|c}
\toprule
  & \multicolumn{4}{c}{\textbf{Jigsaw to Shelf}} & \multicolumn{4}{c}{\textbf{Clothes Iron to Shelf}} & \multicolumn{4}{c}{\textbf{Shoe to Shoerack}} \\
  \cmidrule(lr){2-5} \cmidrule(lr){6-9} \cmidrule(lr){10-13} 
  & \multicolumn{3}{c}{\textbf{Tier 1}} & \textbf{Tier 2} & \multicolumn{3}{c}{\textbf{Tier 1}} & \textbf{Tier 2} & \multicolumn{3}{c}{\textbf{Tier 1}} & \textbf{Tier 2} \\
  \cmidrule(lr){2-4} \cmidrule(lr){5-5} \cmidrule(lr){6-8} \cmidrule(lr){9-9} \cmidrule(lr){10-12} \cmidrule(lr){13-13}
   & \textbf{No Depth} & \textbf{No DINO} & \textbf{POGS} & \textbf{POGS} & \textbf{No Depth} & \textbf{No DINO} & \textbf{POGS} & \textbf{POGS} & \textbf{No Depth} & \textbf{No DINO} & \textbf{POGS} & \textbf{POGS} \\
\midrule
\textbf{Max Consecutive OR} & 1 & 0 & \textbf{11} & 3 & 3 & 0 & \textbf{12} & 6 & 2 & 0 & \textbf{8} & 7 \\
\textbf{Mean Consecutive OR} & 0.33 & 0 & \textbf{4.4} & 1.6 & 0.6 & 0 & \textbf{7.2} & 3 & 1.0 & 0 & \textbf{6.4} & 3.8 \\
\textbf{Successful Pick Rate} & 2/4 & 0/3 & \textbf{23/27} & 9/16 & 6/9 & 0/3 & \textbf{32/36} & 15/18 & 7/10 & 0/3 & \textbf{34/38} & 20/24 \\
\textbf{Successful Place Rate} & 1/2 & - & \textbf{22/23} & 8/9 & 3/6 & - & \textbf{28/32} & 13/15 & 5/7 & - & \textbf{32/34} & 19/20 \\
\textbf{Mean Position Error (cm)} & 2.1 & - & \textbf{2.5} & 1.8 & 7.0 & - & \textbf{3.4} & 2.2 & 4.7 & - & \textbf{4.1} & 3.5 \\
\textbf{Std Position Error (cm)} & 0.0 & - & \textbf{1.0} & 1.1 & 3.7 & - & \textbf{2.6} & 0.8 & 1.5 & - & \textbf{1.5} & 1.3 \\

\bottomrule
\end{tabular}
}
\caption{\textbf{Object Reset Results} \textit{Consecutive OR} refers to a single trial with repeated Object Resets without losing tracking out of five trials for main experiments, and three trials for ablations. Note that the denominators for the successful pick rate and successful place rate metrics vary across trials. This variation arises because each trial was executed until a grasping failure occurred—i.e., when the error in object state estimation became too high to recover—resulting in a different total number of reset attempts per trial.}
\label{table:sequential}
\vspace{-0.5em}
\end{table*}

\begin{figure}[h]
    \centering
    \includegraphics[width=\linewidth]{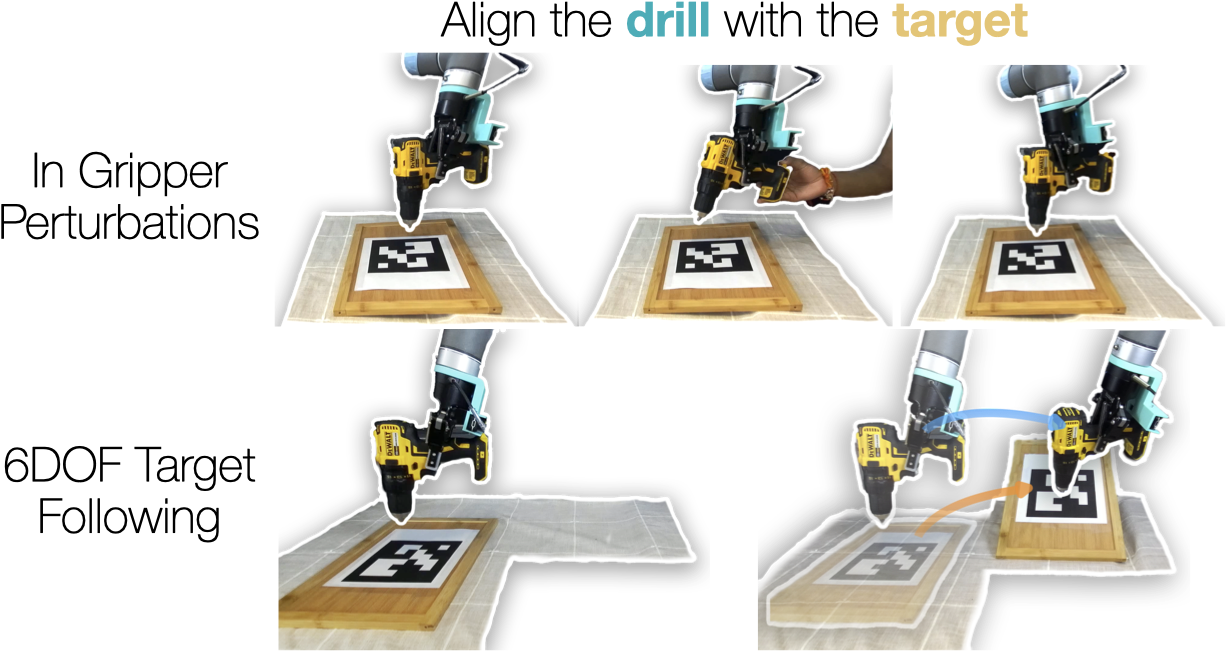}
    \caption{\textbf{Tool Servoing Experimental Setup} The robot continuously attempts to align the tracked tool with the target. \textit{Top:} A human perturbs the tracked tool while in the robot's gripper. The robot adjusts its end-effector position with closed-loop control to re-align the object with the target. \textit{Bottom:} As a human shifts and rotates the target into new poses, the robot moves so the tool follows the target while maintaining alignment.}
    \label{fig:servoing}
\end{figure}
\begin{table}
\centering
\vspace{0.5em}
{\footnotesize
\begin{tabular}{lcc|cc}
\toprule
   & \multicolumn{2}{c}{\textbf{Tier 1}} & \multicolumn{2}{c}{\textbf{Tier 2}} \\
  \cmidrule(lr){2-3} \cmidrule(lr){4-5} 
   \textbf{Perturbations} & \textbf{Success Rate} & \textbf{Time (s)} & \textbf{Success Rate} & \textbf{Time (s)} \\
\midrule
\textbf{Clockwise} & 24/25 & 6.30 & 20/25 & 12.26 \\
\textbf{CCW} & 24/25 & 5.72 & 20/25 & 13.06 \\
\textbf{Follow Target} & 24/25 & - & 21/25 & - \\
\bottomrule
\end{tabular}
}
\caption{\textbf{Tool Servoing Results} We record results across 5 trials for each tier where each trial has the target object move to 5 poses and at each pose, the tool is perturbed clockwise and counter clockwise by 15-30 \degree. For the perturbations, we measure the time it takes for the drill to recover from the perturbation and realign with the target. Since the target was constantly moving over time, recovery time wasn't recorded for the follow target experiment.}
\label{tab:servoing}
\vspace{-0.7em}

\end{table}

\section{Physical Experiments}
For physical experiments, we use a UR5 robotic arm with a static ZED 2 stereo camera. The \algabbr model is trained and initialized on a PC workstation with an NVIDIA 4090 GPU. We evaluate \algabbr on two robotic manipulation tasks across various objects. These tasks test \algabbr's ability to track objects of interest when manipulated by a robot or a human. 
% The first task is a sequential pick and place where the robot puts a queried object onto a queried target object, then both objects are randomly reconfigured to different poses by a human, and the robot has to perform the pick-and-place again continuously until tracking failure. The second task is a tool-in-gripper visual servoing task where the robot grasps a tool and points it towards the target as the target moves or as the tool is adjusted in grasp to demonstrate robustness to grasp perturbations.

Both tasks begin with the UR5 using a wrist-mounted ZED-Mini stereo camera to scan a scene and initialize a \algabbr. Scene scanning with the predefined hemispherical trajectory takes on average 2 minutes and training a \algabbr takes on average 3 minutes.

\subsection{Sequential Object Reset}
This experiment evaluates \algabbr's localization accuracy in sequential object reset tasks guided by natural language. The tasks involve irregular objects of various shapes, sizes, and weights: a jigsaw, clothes iron, shoe, shelf, and shoebox.

Before each trial, a human randomly perturbs the positions and orientations of all objects, with perturbations defined in two tiers: In tier 1, objects could be translated anywhere within the ZED 2 camera frustum but rotated only up to $90\degree$ around the vertical axis from their initial configuration. In tier 2, objects could be translated anywhere within the frustum and rotated to any magnitude around the vertical axis.

The operator provides natural language instructions specifying which object to grasp and where to place it. The robot executes the planned grasp on the target object, adjusts its orientation in the gripper to align with the major axes of the target placement object, and moves the grasped object to the placement location. After each pick-and-place operation, the scene is reset by placing the grasp object back onto the tabletop, after which the human operator perturbs the objects. Tracking remains running the entire time, and these consecutive object resets continue until POGS loses tracking of the objects, defined as when repeated grasp planning failures occur due to irrecoverable errors in object state estimation.

We assess and report in Table \ref{table:sequential} the performance across 3 pick objects and 2 place objects, conducting five trials per pick object on both tiers. The performance metrics included the maximum and mean number of consecutive successful object resets without losing tracking, the successful object reset rates, and the mean and standard deviation of the translation error between the intended and actual placement positions. For example, in the "Clothes Iron to Shelf" task under Tier 1, POGS achieved a maximum of 12 consecutive successful object resets, with a successful pick rate of 32 out of 36 attempts and a mean translation error of 3.4 cm measured by calipers on reference markers made to each object. Under Tier 2, despite more extreme perturbations including full object rotations, POGS achieves 6 consecutive operations, and a mean translation error of 2.2 cm. Similar performance trends were observed in the other tasks, where POGS consistently outperformed ablations that either had depth perception turned off or were optimized with RGB substituting for DINO features. The ablations highlight the critical role that both depth perception and robust visual features play in achieving accurate object localization and successful sequential object resets. 

\subsection{Tool Servoing}
% \textcolor{blue}{Kush: Mentioned in notes that this is too detailed. Should I add some of this detail to the table caption. Worried that if I take any of it out the experiment method/eval won't be clear, but happy to cut stuff out/reword}
We consider a tabletop workspace with a tool object and an ArUco marker fixed to a target object surface. In these experiments, the tool object is a drill and the target object is a wooden platform. The tool is manually annotated with a coordinate frame at the tool tip to indicate which component to align with the target (i.e. align the drill tip perpendicularly). The robot then grasps the tool and aligns it to the tracked ArUco marker. The robot then performs 6DoF visual servoing such that the tool tip remains in its relative target pose to the object. We leave tool rotation about its target axis (i.e drill bit axis) unconstrained, and pick the orientation during servoing which minimizes tool motion. During each trial, the target object is moved to five random poses in the workspace and we record the success rate for how often the tool follows the object. At each pose, a human moves the drill 15-30\degree\xspace clockwise or counterclockwise about the grasp axis, and records how often the robot can adapt to this perturbation and locate the drill tip within 3cm and $5^\circ$ of the target. We also report the average time taken for the robot to adapt to the perturbed gripper pose.

We run this experiment on two tiers with five trials each tier. Tier 1 experiments have the target object stay on the tabletop plane and can move anywhere within a 55 cm by 50 cm square, and the tool orientation in-grasp can be changed up to 15 degrees. Tier 2 experiments have the target moving in 3D space where it can move anywhere within a 55 cm by 72 cm by 10 cm box such that the ArUco marker was visible to the camera and the robot joints did not occlude the ArUco marker from the camera. 

The results are reported in Table~\ref{tab:servoing}. Overall, \algabbr can be used to recover from tool perturbances in gripper up to $15\degree$ in 48 of 50 trials at an average of 6.01 seconds. When the tool-in-gripper rotation increased to $30\degree$, the success rate drops to 40 of 50 trials at a longer time average of 12.66 seconds largely because higher object deltas are harder to track. Similarly, for 2D space, the tool was able to follow the target object 24 of 25 trials but in 3D space that went down to 21 of 25 trials.   

\section{Limitations}

One key limitation of this work is that the online tracking frequency is limited to 5Hz on an NVIDIA 4090 GPU due to computational bottlenecks. This includes approximately 140ms latency for DINO feature extraction using a ViT-S, and multiple steps of optimization necessary per frame iteration to adjust the pose parameters of each object until convergence. As a result, objects had to be moved slowly with no sudden motions or quick changes in direction to avoid losing track of them. In future work, we will parallelize depth inference and DINO feature extraction for tracking speed optimization.

Another limitation is that objects that are partially occluded (by a hand, a robot gripper, etc.) have less robust tracking compared to fully unobstructed objects due to degraded tracking feature alignment between the real features and rendered features. In future work, we will add an end-effector regularization term for objects grasped by the robot as this is a useful prior constraining where the object can be in the workspace. Furthermore, we can also develop robot gripper masking to increase the alignment between rendered features and real features.
\section{Conclusion}

% In this work, we introduce Persistent Object Gaussian Splat (POGS), a system that distills language, grouping, and self-supervised visual features into a 3D Gaussian Splat. 
% With language and grouping features, we are able to query and localize objects with natural language and with self-supervised visual features, we are able to track multiple objects persistently by updating the 3DGS representation online. Results suggest that \algabbr is able to perform robotics tasks such as language-conditioned sequential pick-and-place up to 12 times in a row for a given object and servo tool-in-gripper with grasped tool perturbations 80\% of the time with perturbations of 30\degree.
In this work, we present Persistent Object Gaussian Splat (POGS), a system for tracking and manipulating irregularly shaped, previously unseen objects in dynamic environments. By integrating language, grouping, and self-supervised visual features into an explicit 3D Gaussian representation, POGS aims to address some of the challenges associated with CAD-based, NeRF-based, and conventional point cloud methods. Our experimental results suggest that POGS can maintain object state estimates during tasks such as object resets and tool servoing.

\renewcommand*{\bibfont}{\footnotesize}
\printbibliography

\end{document}